\documentclass[12pt]{article}
\usepackage{graphicx,color}
\usepackage{amsfonts}
\usepackage{amssymb}
\usepackage{amsmath} 
\usepackage{amsthm}
\usepackage{algorithm}
\usepackage{algorithmic}
\usepackage[margin=1.3in]{geometry}

\title{Clustered Multi-Task Learning:\\
a Convex Formulation}

\author{\small\bf Laurent Jacob\thanks{To whom correspondance should
    be addressed: \small 35, rue Saint Honor\'e, F-77300 Fontainebleau, France.}\\
\small Mines ParisTech, CBIO\\
\small Institut Curie, Paris, F-75248 France\\
\small INSERM, U900, Paris, F-75248 France\\
\small \texttt{laurent.jacob@mines-paristech.fr} \\
\and
\small\bf Francis Bach\\
\small INRIA -- WILLOW Project Team\\
\small \'Ecole Normale Sup\'erieure, DI\\
\small (CNRS/ENS/INRIA UMR 8548)\\
\small \texttt{francis.bach@mines.org} \\
\and
\small\bf Jean-Philippe Vert \\
\small Mines ParisTech, CBIO\\
\small Institut Curie, Paris, F-75248 France\\
\small INSERM, U900, Paris, F-75248 France\\
\small \texttt{jean-philippe.vert@mines-paristech.fr} \\
}

\newcommand{\RR}{\mathbb{R}}
\newcommand{\X}{\mathcal{X}}
\newcommand{\Y}{\mathcal{Y}}
\newcommand{\Ical}{\mathcal{I}}
\newcommand{\Jcal}{\mathcal{J}}
\newcommand{\Mcal}{\mathcal{M}}
\newcommand{\LG}{\mathcal{L}}
\newcommand{\ORT}{\mathcal{O}}
\newcommand{\Scal}{\mathcal{S}}
\newcommand{\tr}{{\rm tr}}
\newcommand{\diag}{\mathop{\rm diag}}
\newcommand{\1}{{\bf 1}}
\newcommand{\E}{\varepsilon}
\newcommand{\lC}{\ell_c}
\newcommand{\SC}{\Sigma_c}

\begin{document}

\maketitle

\begin{abstract}
  In multi-task learning several related tasks are considered
  simultaneously, with the hope that by an appropriate sharing of
  information across tasks, each task may benefit from the others. In
  the context of learning linear functions for supervised
  classification or regression, this can be achieved by including a
  priori information about the weight vectors associated with the
  tasks, and how they are expected to be related to each other. In
  this paper, we assume that tasks are clustered into groups, which are
  unknown beforehand, and that tasks within a group have similar
  weight vectors. We design a new spectral norm that encodes this a
  priori assumption, without the prior knowledge of the partition of
  tasks into groups, resulting in a new convex optimization
  formulation for multi-task learning. We show in simulations on
  synthetic examples and on the \textsc{iedb} MHC-I binding dataset,
  that our approach outperforms well-known convex methods for
  multi-task learning, as well as related non convex methods dedicated
  to the same problem.
\end{abstract}

 \section{Introduction}

 Regularization has emerged as a dominant theme in machine learning
 and statistics, providing an intuitive and principled tool for
 learning from high-dimensional data. In particular, regularization by
 squared Euclidean norms or squared Hilbert norms has been thoroughly
 studied in various settings, leading to efficient practical
 algorithms based on linear algebra, and to very good theoretical
 understanding (see, e.g.,
 \cite{Wahba1990Spline,Girosi1995Regularization}).
 In recent years, regularization by non Hilbert norms, such as
 $\ell^p$ norms with $p\neq 2$, has also generated considerable
 interest for the inference of linear functions in supervised
 classification or regression. Indeed, such norms can sometimes both
 make the problem statistically and numerically better-behaved, and
 impose various a priori knowledge on the problem. For example, the
 $\ell^1$-norm (the sum of absolute values) imposes some of the
 components to be equal to zero and is widely used to estimate sparse
 functions~\cite{Tibshirani1996Regression}, while
 various combinations of $\ell^p$ norms can be defined to impose
 various sparsity patterns.
 
  While most recent work has focused on studying the properties of
 simple well-known norms, we take the opposite approach in this paper.
 That is, assuming a given prior knowledge, how can we design a norm
 that will enforce it?

 More precisely, we consider the problem of multi-task learning, which
 has recently emerged as a very promising research direction for
 various applications \cite{Bakker2003Task}. In multi-task learning
 several related inference tasks are considered simultaneously, with
 the hope that by an appropriate sharing of information across tasks,
 each one may benefit from the others. When linear functions are
 estimated, each task is associated with a weight vector, and a common
 strategy to design multi-task learning algorithm is to translate some
 prior hypothesis about how the tasks are related to each other into
 constraints on the different weight vectors. For example, such
 constraints are typically that the weight vectors of the different
 tasks belong (a) to a Euclidean ball centered at the
 origin~\cite{Evgeniou2005Learning}, which implies no sharing of
 information between tasks apart from the size of the different
 vectors, i.e., the amount of regularization, (b) to a ball of unknown
 center~\cite{Evgeniou2005Learning}, which enforces a similarity
 between the different weight vectors, or (c) to an unknown
 low-dimensional
 subspace~\cite{Abernethy2006Low-rank,Argyriou2007Multi-task}.

 In this paper, we consider a different prior hypothesis that we
 believe could be more relevant in some applications: the hypothesis
 that \emph{the different tasks are in fact clustered into different
   groups, and that the weight vectors of tasks within a group are
   similar to each other}. A key difference with
 \cite{Evgeniou2005Learning}, where a similar hypothesis is studied,
 is that we don't assume that the groups are known a priori, and in a
 sense our goal is both to identify the clusters and to use them for
 multi-task learning.  An important situation that motivates this
 hypothesis is the case where most of the tasks are indeed related to
 each other, but a few ``outlier'' tasks are very different, in which
 case it may be better to impose similarity or low-dimensional
 constraints only to a subset of the tasks (thus forming a cluster)
 rather than to all tasks. Another situation of interest is when one
 can expect a natural organization of the tasks into clusters, such as
 when one wants to model the preferences of customers and believes
 that there are a few general types of customers with similar
 preferences within each type, although one does not know beforehand
 which customers belong to which types. Besides an improved
 performance if the hypothesis turns out to be correct, we also expect
 this approach to be able to identify the cluster structure among the
 tasks as a by-product of the inference step, e.g., to identify
 outliers or groups of customers, which can be of interest for further
 understanding of the structure of the problem.

 In order to translate this hypothesis into a working algorithm, we
 follow the general strategy mentioned above which is to design a norm
 or a penalty over the set of weights which can be used as
 regularization in classical inference algorithms. We construct such a
 penalty by first assuming that the partition of the tasks into
 clusters is known, similarly to~\cite{Evgeniou2005Learning}. We then
 attempt to optimize the objective function of the inference algorithm
 over the set of partitions, a strategy that has proved useful in
 other contexts such as multiple kernel
 learning~\cite{Lanckriet2004Learning}. This optimization problem over
 the set of partitions being computationally challenging, we propose
 a convex relaxation of the problem which results in an efficient
 algorithm.

 \section{Multi-task learning with clustered tasks}

We consider $m$ related inference tasks that attempt to learn linear
functions over $\X = \RR^d$ from a training set of input/output pairs
$(x_i,y_i)_{i=1,\ldots,n}$, where $x_i \in \X$ and $y_i \in\Y$. In the
case of binary classification we usually take $\Y = \{-1,+1\}$, while
in the case of regression we take $\Y=\RR$. Each training example
$(x_i,y_i)$ is associated to a particular task $t\in[1,m]$, and we
denote by $\Ical(t) \subset [1,n]$ the set of indices of training
examples associated to the task $t$. Our goal is to infer $m$ linear
functions $f_t(x) = w_t^\top x$, for $t=1,\ldots,m$, associated to the
different tasks. We denote by $W = (w_1 \ldots w_m)$ the $d\times m$
matrix whose columns are the successive vectors we want to estimate.

We fix a loss function $l:\RR\times\Y\mapsto\RR$ that quantifies by
$l(f(x),y)$ the cost of predicting $f(x)$ for the input $x$ when the
correct output is $y$. Typical loss functions include the square error
in regression $l(u,y)=\frac{1}{2}(u-y)^2$ or the hinge loss in binary
classification $l(u,y)=\max(0,1-uy)$ with $y \in \{-1,1\}$. The
empirical risk of a set of linear classifiers given in the matrix $W$
is then defined as the average loss over the training set:
\begin{equation}\label{eq:loss}
\ell(W) = \frac{1}{n} \sum_{t=1}^m  \sum_{i\in\Ical(t)} l(w_t^\top x_i , y_i)  \,.
\end{equation}

In the sequel, we will often use the $m \!\times\! 1$ vector $\1$ composed
of ones, the $m\!\times \! m$ projection matrices $U\! =\! \1\1^\top / m$
whose entries are all equal to $1/m$, as well as the projection matrix
$\Pi\!= \! I- U$.

In order to learn simultaneously the $m$ tasks, we follow the now
well-established approach which looks for a set of weight vectors $W$
that minimizes the empirical risk regularized by a penalty functional,
i.e., we consider the problem:
\begin{equation}\label{eq:pb1}
\min_{W\in\RR^{d\times m}} \ell(W)+ \lambda \Omega(W)\,,
\end{equation}
where $\Omega(W)$ can be designed from prior knowledge to constrain
some sharing of information between tasks. For example,
\cite{Evgeniou2005Learning} suggests to penalize both the norms of the
$w_i$'s and their variance, i.e., to consider a function of the form:
\begin{equation}\label{eq:variancenorm}
\Omega_{variance}(W) = \|\bar{w}\|^2 + \frac{\beta}{m} \sum_{i=1}^m \| w_i - \bar{w}\|^2\,,
\end{equation}
where $\bar{w} = \left(\sum_{i=1}^n w_i\right)/m$ is the mean weight
vector. This penalty enforces a clustering of the $w_i's$ towards
their mean when $\beta$ increases.
Alternatively,~\cite{Argyriou2007Multi-task} propose to penalize the
trace norm of $W$:
\begin{equation}
\label{eq:tracenorm}
\Omega_{trace} (W) = \sum_{i=1}^{\min(d,m)}\sigma_i(W)\,,
\end{equation}
where $\sigma_1(W),\ldots,\sigma_{\min(d,m)}(W)$ are the successive
singular values of $W$. This enforces a low-rank solution in $W$,
i.e., constrains the different $w_i$'s to live in a low-dimensional
subspace.

Here we would like to define a penalty function $\Omega(W)$ that
encodes as prior knowledge that tasks are clustered into $r<m$ groups.
To do so, let us first assume that we know beforehand the clusters,
i.e., we have a partition of the set of tasks into $r$ groups. In that
case we can follow an approach proposed by \cite{Evgeniou2005Learning}
which for clarity we rephrase with our notations and slightly
generalize now. For a given cluster $c\in[1,r]$, let us denote
$\Jcal(c)\subset [1,m]$ the set of tasks in $c$, $m_c = |\Jcal(c)|$
the number of tasks in the cluster $c$, and $E$ the $m\times r$ binary
matrix which describes the cluster assignment for the $m$ tasks, i.e.,
$E_{ij}=1$ if task $i$ is in cluster $j$, $0$ otherwise. Let us
further denote by $\bar{w}_c =  (\sum_{i\in\Jcal(c)} w_i  )/
m_c$ the average weight vector for the tasks in $c$, and recall that
$\bar{w} = \left( \sum_{i=1}^m w_i \right) / m$ denotes the average
weight vector over all tasks. Finally it will be convenient to
introduce the matrix $M=E(E^\top E)^{-1}E^\top$. $M$ can also be
written $L-I$, where $L$ is the normalized Laplacian of the graph $G$
whose nodes are the tasks connected by an edge if and only if they are
in the same cluster. Then we can define three semi-norms of interest
on $W$ that quantify different orthogonal aspects:
\begin{itemize}
\item A global penalty, which measures on average how large the weight vectors are:
\[
\Omega_{mean}(W) = n\|\bar{w}\|^2 = \tr W U W^\top\,.
\]
\item A measure of between-cluster variance, which quantifies how close to each other the different clusters are:
\[
\Omega_{between}(W) = \sum_{c=1}^r m_c \| \bar{w}_c - \bar{w} \|^2 = \tr W (M-U) W^\top.
\]
\item A measure of within-cluster variance, which quantifies the compactness of the different clusters:
\[
\Omega_{within}(W) = \sum_{c=1}^r \left\{ \sum_{i\in\Jcal(c)} \| w_i - \bar{w}_c\|^2\right\} = \tr W (I-M) W^\top \,.
\]
\end{itemize}
We note that both $\Omega_{between}(W)$ and $\Omega_{within}(W)$
depend on the particular choice of clusters $E$, or equivalently of
$M$. We now propose to consider the following general penalty
function:
\begin{equation}\label{eq:omega}
\Omega(W) = \E_M \Omega_{mean}(W) + \E_B \Omega_{between}(W) + \E_W \Omega_{within}(W)\,,
\end{equation}
where $\E_M , \E_B$ and $\E_W$ are three non-negative parameters that
can balance the importance of the different components of the penalty.
Plugging this quadratic penalty into \eqref{eq:pb1} leads to the
general optimization problem:
\begin{equation}\label{eq:pb2}
\min_{W\in\RR^{d\times m}} \ell(W)+ \lambda \tr W \Sigma(M)^{-1} W^\top\,,
\end{equation}
where
\begin{equation}\label{eq:sigmainv}
\Sigma(M)^{-1} = \E_M U + \E_B (M-U) + \E_W (I-M)\,.
\end{equation}
Here we use the notation $\Sigma(M)$ to insist on the fact that this
quadratic penalty depends on the cluster structure through the matrix
$M$. Observing that the matrices $U$, $M-U$ and $I-M$ are orthogonal projections onto orthogonal supplementary subspaces, we easily get from \eqref{eq:sigmainv}: 
\begin{equation}\label{eq:sigma}
\Sigma(M) =  \E_M^{-1} U + \E_B^{-1} (M-U) + \E_W^{-1} (I-M) =  \E_W^{-1} I + ( \E_M^{-1} -  \E_B^{-1}) U + ( \E_B^{-1} -  \E_W^{-1})M\,.
\end{equation}

By choosing particular values for $\E_M, \E_B$ and $\E_W$ we can
recover several situations, In particular:
\begin{itemize}
\item For $\E_W = \E_B = \E_M = \E $, we simply recover the Frobenius norm of $W$, which does not put any constraint on the relationship between the different tasks:
\[
\Omega(W) = \E \tr W W^\top = \E \sum_{i=1}^m \|w_i\|^2\,.
\]
\item For $\E_W = \E_B > \E_M $, we recover the penalty of \cite{Evgeniou2005Learning} without clusters:
\[
\Omega(W) = \tr W\left(\E_M U + \E_B(I-U)\right)W^\top = \E_M n \|\bar{w}\|^2 + \E_B \sum_{i=1}^m \|w_i - \bar{w}\|^2\,.
\]
In that case, a global similarity between tasks is enforced, in addition to the general constraint on their mean. The structure in clusters plays no role since the sum of
the between- and within-cluster variance is independent of the
particular choice of clusters.
\item For $\E_W > \E_B = \E_M$ we recover the penalty of \cite{Evgeniou2005Learning} with clusters:
\begin{align}
\notag \Omega(W) &= \tr W\left(\E_M M + \E_W(I-M)\right)W^\top \\
&= \E_M \sum_{c=1}^r  
\left\{ m_c \|\bar{w}_c \|^2 + \frac{\E_W}{\E_M} \sum_{i \in \Jcal(c)} \|w_i - \bar{w}_c\|^2 \right\} \,.
\end{align}
\end{itemize}
In order to enforce a cluster hypothesis on the tasks, we therefore
see that a natural choice is to take $\E_W > \E_B > \E_M$ in
\eqref{eq:omega}. This would have the effect of penalizing more the
within-cluster variance than the between-cluster variance, hence
promoting compact clusters. Of course, a major limitation at this
point is that we assumed the cluster structure known a priori (through
the matrix $E$, or equivalently $M$). In many cases of interest, we
would like instead to learn the cluster structure itself from the
data. We propose to learn the cluster structure in our framework by
optimizing our objective function \eqref{eq:pb2} both in $W$ and $M$,
i.e., to consider the problem:
\begin{equation}\label{eq:pb23}
  \min_{W\in\RR^{d\times m}, M \in \Mcal_r} \ell(W)+ \lambda \tr W \Sigma(M)^{-1} W^\top\,,
\end{equation}
where $\Mcal_r$ denotes the set of matrices $M=E(E^\top E)^{-1}E^\top$
defined by a clustering of the $m$ tasks into $r$ clusters and
$\Sigma(M)$ is defined in \eqref{eq:sigma}. Denoting by $\Scal_r =
\left\{\Sigma(M) : M\in\Mcal_r \right\}$ the corresponding set of
positive semidefinite matrices, we can equivalently rewrite the
problem as:
\begin{equation}\label{eq:pb3}
\min_{W\in\RR^{d\times m}, \Sigma \in \Scal_r} \ell(W)+ \lambda \tr W \Sigma^{-1} W^\top\,.
\end{equation}
The objective function in
\eqref{eq:pb3} is jointly convex in $W\in\RR^{d\times m}$ and
$\Sigma\in\Scal_{+}^m$, the set of $m\times m$ positive semidefinite
matrices, however the (finite) set $\Scal_r$ is not convex, making
this problem intractable. We are now going to propose a convex
relaxation of \eqref{eq:pb3} by optimizing over a convex set of
positive semidefinite matrices that contains $\Scal_r$.

\section{Convex relaxation}

In order to formulate a convex relaxation of~\eqref{eq:pb3}, let us
first observe that in the penalty term~\eqref{eq:omega} the cluster
structure only contributes to the second and third terms
$\Omega_{between}(W)$ and $\Omega_{within}(W)$, and that these
penalties only depend on the centered version of $W$. In terms of
matrices, only the last two terms of $\Sigma(M)^{-1}$
in~\eqref{eq:sigmainv} depend on $M$, \emph{i.e.}, on the clustering,
and these terms can be re-written as:
\begin{equation}\label{lemdefrancis}
\E_B (M-U) + \E_W (I-M) =  \Pi( \E_B M  +  \E_W (I-M) )\Pi  .
\end{equation}
Indeed, it is easy to check that $M - U = M\Pi = \Pi M \Pi $, and that
$I-M=I-U-(M-U) = \Pi - \Pi M \Pi = \Pi(I-M)\Pi$. Intuitively,
multiplying by $\Pi$ on the right (\emph{resp.} on the left) centers
the rows (\emph{resp.} the columns) of a matrix, and both $M-U$ and
$I-M$ are row- and column-centered.

To simplify notations, let us introduce $\widetilde{M}= \Pi M \Pi $.
Plugging~\eqref{lemdefrancis} in~\eqref{eq:sigmainv}
and~\eqref{eq:pb23}, we get the penalty
\begin{equation}\label{eq:newpenalty}
\tr W \Sigma(M)^{-1} W^\top = \E_M  \left(\tr W^\top W U \right)
+   (W\Pi)( \E_B \widetilde{M}  +  \E_W (I-\widetilde{M} ) )  (W \Pi) ^\top,
\end{equation}
in which, again, only the second part needs to be optimized with
respect to the clustering $M$. Denoting $\Sigma_c^{-1}(M) = \E_B \widetilde{M} +
\E_W (I-\widetilde{M} )$, one can express $\Sigma_c(M)$, using the fact
that $\widetilde{M}$ is a projection:
\begin{equation}
\label{eq:SS}
\Sigma_c(M) = \left(\E_B^{-1} - \E_W^{-1} \right)\widetilde{M}  + \E_W^{-1}I.
\end{equation}
$\Sigma_c$ is characterized by $\widetilde{M} = \Pi  M \Pi $, that is discrete by construction,
hence the non-convexity of $\Scal_r$. We have the natural constraints
$M\geq 0$ (i.e., $\widetilde{M} \geq - U$), $0\preceq M\preceq I$ (i.e., $0\preceq \widetilde{M} \preceq \Pi$ and $\tr M = r$ (i.e., $\tr \widetilde{M} = r -1 $). A possible convex
relaxation of the discrete set of matrices $\widetilde{M}$  is therefore $\{\widetilde{M}: 0\preceq
\widetilde{M}\preceq I,\;\tr \widetilde{M}= r-1\}$. This gives an equivalent convex set $\Scal_c$ for $\Sigma_c$, namely:
\begin{equation}\label{eq:Sc}
\Scal_c = \left\{\Sigma_c \in \Scal_+^m : \alpha I \preceq \Sigma \preceq
  \beta I, \tr \Sigma = \gamma \right\} \,,
  \end{equation}
  with $\alpha = \E_W^{-1}$, $\beta=\E_B^{-1}$ and
  $\gamma=(m-r+1)\E_W^{-1} + (r-1)\E_B^{-1}$. Incorporating the first
  part of the penalty~\eqref{eq:newpenalty} into the empirical risk
  term by defining $\ell_c(W) = \lambda\ell(W) + \E_M \left(\tr W^\top
    W U \right)$, we are now ready to state our relaxation of~\eqref{eq:pb3}:
\begin{equation}\label{eq:relaxation}
\min_{W\in\RR^{d\times m}, \Sigma_c \in \Scal_c} \ell_c(W)+ \lambda \tr \Pi W \Sigma_c^{-1} W^\top \Pi\,.
\end{equation}

\subsection{Reinterpretation in terms of norms}

We denote $\|W\|_c^2 = \min_{\SC\in \Scal_c}\tr W\SC^{-1}W^T$ the
\emph{cluster norm} (CN). For any convex set $\Scal_c$, we obtain a
norm on $W$ (that we apply here to its centered version). By putting
some different constraints on the set $\Scal_c$, we obtain different
norms on $W$, and in fact all previous multi-task formulations may be
cast in this way, \emph{i.e.}, by choosing a specific set of positive
matrices $\Scal_c$ (\emph{e.g.}, trace constraint for the trace norm,
and simply a singleton for the Frobenius norm).  Thus, designing norms
for multi-task learning is equivalent to designing a set of positive
matrices. In this paper, we have investigated a specific set adapted
for clustered-tasks, but other sets could be designed in other
situations.

Note that we have selected a simple \emph{spectral} convex set
$\Scal_c$ in order to make the optimization simpler in
Section~\ref{sec:primal}, but we could also add some additional
constraints that encode the point-wise positivity of the matrix $M$.
Finally, when $r=1$ (one clusters) and $r=m$ (one cluster per task),
we get back the formulation of \cite{Evgeniou2005Learning}.

\subsection{Reinterpretation as a convex relaxation of K-means}

In this section we show that the semi-norm $\| \Pi W\|_c^2$ that we
have designed earlier, can be interpreted as a convex relaxation of
K-means on the tasks~\cite{Deodhar2007framework}.  Indeed, given $W
\in \RR^{d \times m} $, K-means aims to decompose it in the form $W =
\mu E^\top$ where $\mu \in \RR^{d \times r} $ are cluster centers and
$E$ represents a partition. Given the partition $E$, the matrix $\mu$
is found by minimizing $ \min_{\mu} \| W^\top - E \mu^\top\|_F^2 $.
Thus, a natural strategy outlined by~\cite{Deodhar2007framework}, is
to alternate between optimizing $\mu$, the partition $E$ and the
weight vectors $W$. We now show that our convex norm is obtained when
minimizing in closed form with respect to $\mu$ and relaxing.

By translation invariance, this is equivalent to minimizing $
\min_{\mu } \| \Pi W^\top - \Pi E \mu^\top \|_F^2 $. If we add a
penalization on $\mu$ of the form $\lambda \tr E^\top E \mu \mu^\top$,
then a short calculation shows that the minimum with respect to $\mu$
(i.e., after optimization of the cluster centers) is equal to
\begin{displaymath}
\tr  \Pi W^\top W \Pi ( \Pi E (E^\top E)^{-1}E^\top \Pi / \lambda + I )^{-1} 
=  \tr  \Pi W^\top W \Pi ( \Pi M \Pi / \lambda + I )^{-1}.
\end{displaymath}
By comparing with Eq.~\eqref{eq:SS}, we see that our formulation is
indeed a convex relaxation of K-means.

\subsection{Primal optimization}
\label{sec:primal}

Let us now show in more details how~\eqref{eq:relaxation} can be
solved efficiently. Whereas a dual formulation could be easily derived
following~\cite{Lanckriet2004Learning}, a direct approach is to
rewrite~\eqref{eq:relaxation} as
\begin{equation}
  \label{eq:itsp}
  \min_{W\in\RR^{d\times m}}\left(\lC(W)+\min_{\SC\in \Scal_c}\tr \Pi W\SC^{-1}W^T \Pi \right)
\end{equation}
which, if $\lC$ is differentiable, can be directly optimized by
gradient-based methods on $W$ since $\| \Pi W\|_c^2 = \min_{\SC\in
  \Scal_c}\tr \Pi W\SC^{-1}W^T \Pi $ is a quadratic semi-norm of $W$.
This regularization term $\tr \Pi W \SC^{-1} W^\top \Pi $ and its
gradient can be computed efficiently using a semi-closed form. Indeed,
since $\SC$ as defined in \eqref{eq:Sc} is a spectral set (i.e., it
does depend only on eigenvalues of covariance matrices), we obtain a
function of the singular values of $\Pi W$ (or equivalently the
eigenvalues of $W^\top \Pi W$):
\begin{equation*}
  \min_{\SC\in \Scal_c}\tr \Pi W \SC^{-1} W^\top \Pi  =
  \min_{\lambda\in\RR^{m},\;\alpha\leq \lambda_i\leq \beta,\;\lambda\1
    = \gamma,\; U\in\ORT^{m}}\tr WU\diag(\lambda)^{-1}U^\top W^\top,
\end{equation*}
where $\ORT^{m}$ is the set of orthogonal matrices in $\RR^{m\times
  m}$. The optimal $U$ is the matrix of the eigenvectors of
$W^\top\Pi W$, and we obtain the value of the objective function at the optimum:
\[
 \min_{\Sigma\in S}\tr \Pi W \Sigma^{-1} W^\top \Pi
=\min_{\lambda\in\RR^{m},\;\alpha\leq \lambda_i\leq \beta,\;\lambda\1
  = \gamma}\sum_{i=1}^m\frac{\sigma_i^2}{\lambda_i},
\]
where $\sigma$ and $\lambda$ are the vectors containing the singular
values of $\Pi W$ and $\Sigma$ respectively. Now, we simply need to be
able to compute this function of the singular values.

The only coupling in this formulation comes from the trace
constraint. The Lagrangian corresponding to this constraint is:
\begin{equation}
  \label{eq:lag}
  \LG(\lambda,\nu) = \sum_{i=1}^m\frac{\sigma_i^2}{\lambda_i} + \nu
  \left(\sum_{i=1}^m\lambda_i - \gamma\right)\,.
\end{equation}
For $\nu\leq 0$, this is a decreasing function of $\lambda_i$, so the
minimum on $\lambda_i \in [\alpha,\beta]$ is reached for $\lambda_i =
\beta$. The dual function is then a linear non-decreasing function of
$\nu$ (since $\alpha\leq\gamma/m\leq\beta$ from the definition of
$\alpha,\beta,\gamma$ in~\eqref{eq:Sc}, which reaches it maximum value
(on $\nu\leq 0$) at $\nu=0$. Let us therefore now consider the dual
for $\nu \geq 0$.  \eqref{eq:lag} is then a convex function of
$\lambda_i$. Canceling its derivative with respect to $\lambda_i$
gives that the minimum in $\lambda\in\RR$ is reached for
$\lambda_i=\sigma_i / \sqrt{\nu}$. Now this may not be in the
constraint set $\left(\alpha,\beta\right)$, so if
$\sigma_i<\alpha\sqrt{\nu}$ then the minimum in
$\lambda_i\in[\alpha,\beta]$ of~\eqref{eq:lag} is reached for
$\lambda_i=\alpha$, and if $\sigma_i>\beta\sqrt{\nu}$ it is reached
for $\lambda_i=\beta$. Otherwise, it is reached for
$\lambda_i=\sigma_i/\sqrt{\nu}$.  Reporting this in~\eqref{eq:lag},
the dual problem is therefore
\begin{equation}
  \label{eq:dual}
 \max_{\nu\geq 0}
  \sum_{i,
    \alpha\sqrt{\nu}\leq\sigma_i\leq\beta\sqrt{\nu}}2\sigma_i\sqrt{\nu} 
+ \sum_{i,
    \sigma_i<\alpha\sqrt{\nu}}\left(\frac{\sigma_i^2}{\alpha}+\nu\alpha\right) 
+ \sum_{i,
    \beta\sqrt{\nu}<\sigma_i}\left(\frac{\sigma_i^2}{\beta}+\nu\beta\right)
-\nu\gamma \,.
\end{equation}

Since a closed form for this expression is known for each fixed value
of $\nu$, one can obtain $\|\Pi W\|_c^2$ (and the eigenvalues of
$\Sigma^*$) by Algorithm \ref{alg:norm}.
\begin{algorithm}
\caption{Computing $\|A\|_c^2$}
\label{alg:norm}
\begin{algorithmic}
\REQUIRE $A,\alpha,\beta,\gamma$.
\ENSURE $\|A\|_c^2$, $\lambda^*$.
\STATE Compute the singular values $\sigma_i$ of $A$.
\STATE Order the
$\frac{\sigma_i^2}{\alpha^2},\frac{\sigma_i^2}{\beta^2}$ in a vector
$I$ (with an additional $0$ at the beginning).
\FORALL{interval $\left(a,b\right)$ of $I$}
\IF{$\frac{\partial\LG(\lambda^*,\nu)}{\partial\nu}$ is canceled on
  $\nu\in\left(a,b\right)$}
\STATE Replace $\nu^*$ in the dual function $\LG(\lambda^*,\nu)$ to get $\|A\|^2_c$, compute
$\lambda^*$ on $\left(a,b\right)$.
\STATE \textbf{return} $\|A\|_c^2$, $\lambda^*$.
\ENDIF
\ENDFOR
\end{algorithmic}
\end{algorithm}
The cancellation condition in Algorithm~\ref{alg:norm} is that the
value canceling the derivative belongs to $\left(a,b\right)$,
\emph{i.e.},
\[
 \nu=\left(\frac{\sum_{i,
      \alpha\sqrt{\nu}\leq\sigma_i\leq\beta\sqrt{\nu}}\sigma_i}{\gamma
    - (\alpha n^-+\beta n^+)}\right)^2\in\left(a,b\right),
\]
where $n^-$ and $n^+$ are the number of $\sigma_i<\alpha\sqrt{\nu}$
and $\sigma_i>\beta\sqrt{\nu}$ respectively. In order to perform the
gradient descent, we also need to compute $\frac{\partial\|\Pi
  W\|^2_c}{\partial W}$. This can be computed directly using
$\lambda^*$, by:
\begin{displaymath}
  \forall i, \frac{\partial\|\Pi W\|^2_c}{\partial
  \sigma_i} = \frac{2\sigma_i}{\lambda_i^*} \textrm{ and } \frac{\partial\|\Pi
  W\|^2_c}{\partial W} = \frac{\partial\|\Pi W\|^2_c}{\partial \Pi W}\Pi.
\end{displaymath}

\section{Experiments}

\subsection{Artificial data}

We generated synthetic data consisting of two clusters of two tasks.
The tasks are vectors of $\RR^d,\; d=30$. For each cluster, a center
$\bar{w}_c$ was generated in $\RR^{d-2}$, so that the two clusters be
orthogonal.  More precisely, each $\bar{w}_c$ had $(d-2)/2$ random
features randomly drawn from $\mathcal{N}(0,\sigma_r^2),\;
\sigma_r^2=900$, and $(d-2)/2$ zero features. Then, each tasks $t$ was
computed as $w_t + \bar{w}_c(t)$, where $c(t)$ was the cluster of $t$.
$w_t$ had the same zero feature as its cluster center, and the other
features were drawn from $\mathcal{N}(0,\sigma_c^2),\; \sigma_c^2=16$.
The last two features were non-zero for all the tasks and drawn from
$\mathcal{N}(0,\sigma_c^2)$. For each task, $2000$ points were
generated and a normal noise of variance $\sigma^2_n=150$ was added.

In a first experiment, we compared our cluster norm $\|.\|^2_c$
with the single-task learning given by the Frobenius norm, and with
the trace norm, that corresponds to the assumption that the tasks live
in a low-dimension space. The multi-task kernel approach being a
special case of CN, its performance will always be between the
performance of the single task and the performance of CN.

In a second setting, we compare CN to alternative methods that differ
in the way they learn $\Sigma$:
\begin{itemize}
\item The \emph{True metric} approach, that simply plugs the actual
  clustering in $E$ and optimizes $W$ using this fixed metric. This
  necessitates to know the true clustering \emph{a priori}, and can be
  thought of like a golden standard.
\item The \emph{k-means} approach, that alternates between optimizing
  the tasks in $W$ given the metric $\Sigma$ and re-learning $\Sigma$
  by clustering the tasks $w_i$~\cite{Deodhar2007framework}. The
  clustering is done by a k-means run $3$ times. This is a non convex
  approach, and different initialization of k-means may result in
  different local minima.
\end{itemize}
We also tried one run of CN followed by a run of \emph{True metric}
using the learned $\Sigma$ reprojected in $\Scal_r$ by
rounding,\emph{i.e.}, by performing k-means on the eigenvectors of the
learned $\Sigma$ (\emph{Reprojected} approach), and a run of
\emph{k-means} starting from the relaxed solution (\emph{CNinit}
approach).

Only the first method requires to know the true clustering a
priori, all the other methods can be run without any knowledge of the
clustering structure of the tasks.

Each method was run with different numbers of training points. The
training points were equally separated between the two clusters and
for each cluster, $5/6$th of the points were used for the first task
and $1/6$th for the second, in order to simulate a natural setting
were some tasks have fewer data. We used the $2000$ points of each
task to build $3$ training folds, and the remaining points were used
for testing. We used the mean RMSE across the tasks as a criterion,
and a quadratic loss for $\ell(W)$.

\begin{figure}[h]
\begin{center}
\small
\begin{tabular}{ccc}
  \includegraphics[width=.5\linewidth]{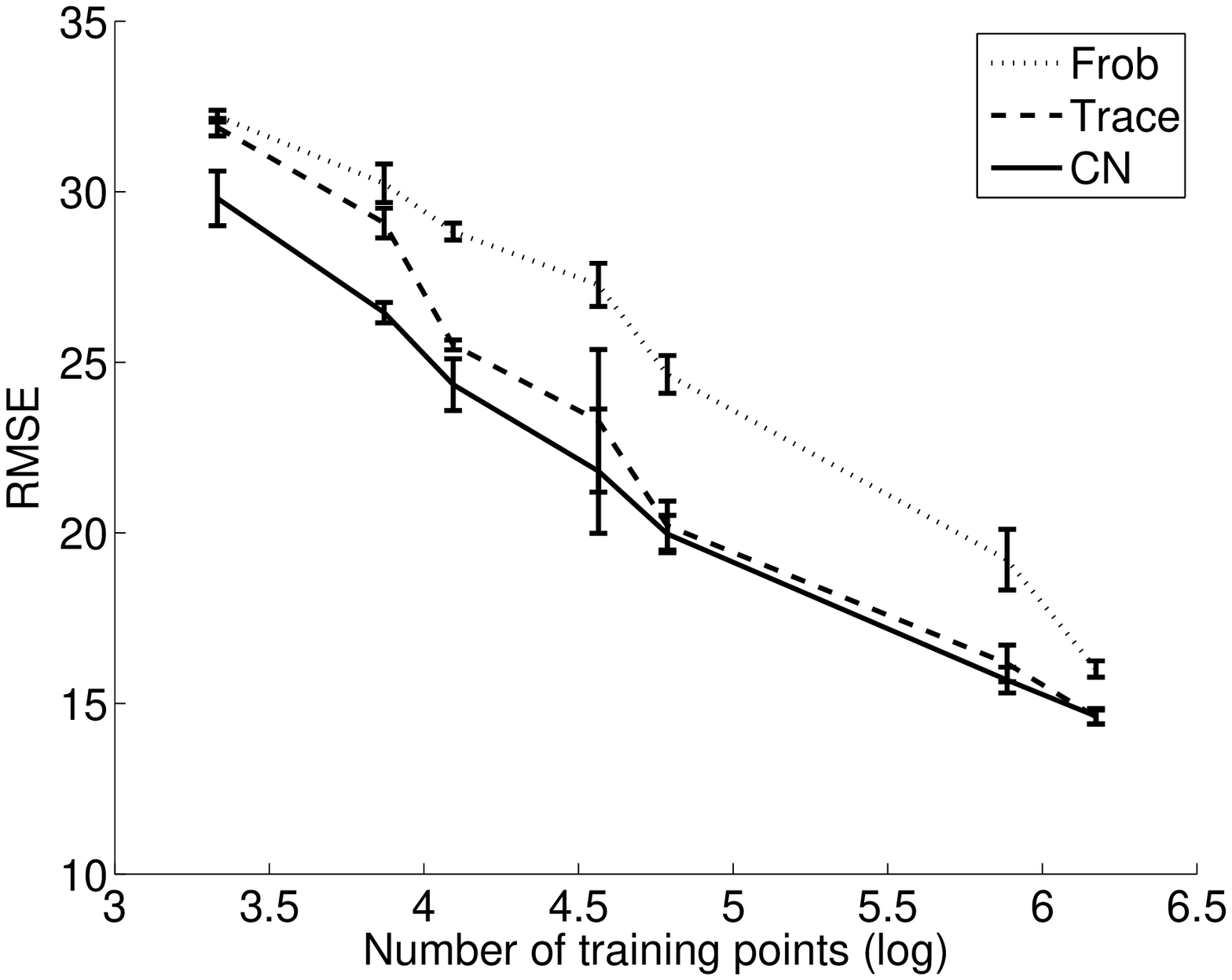}& 
  \includegraphics[width=.5\linewidth]{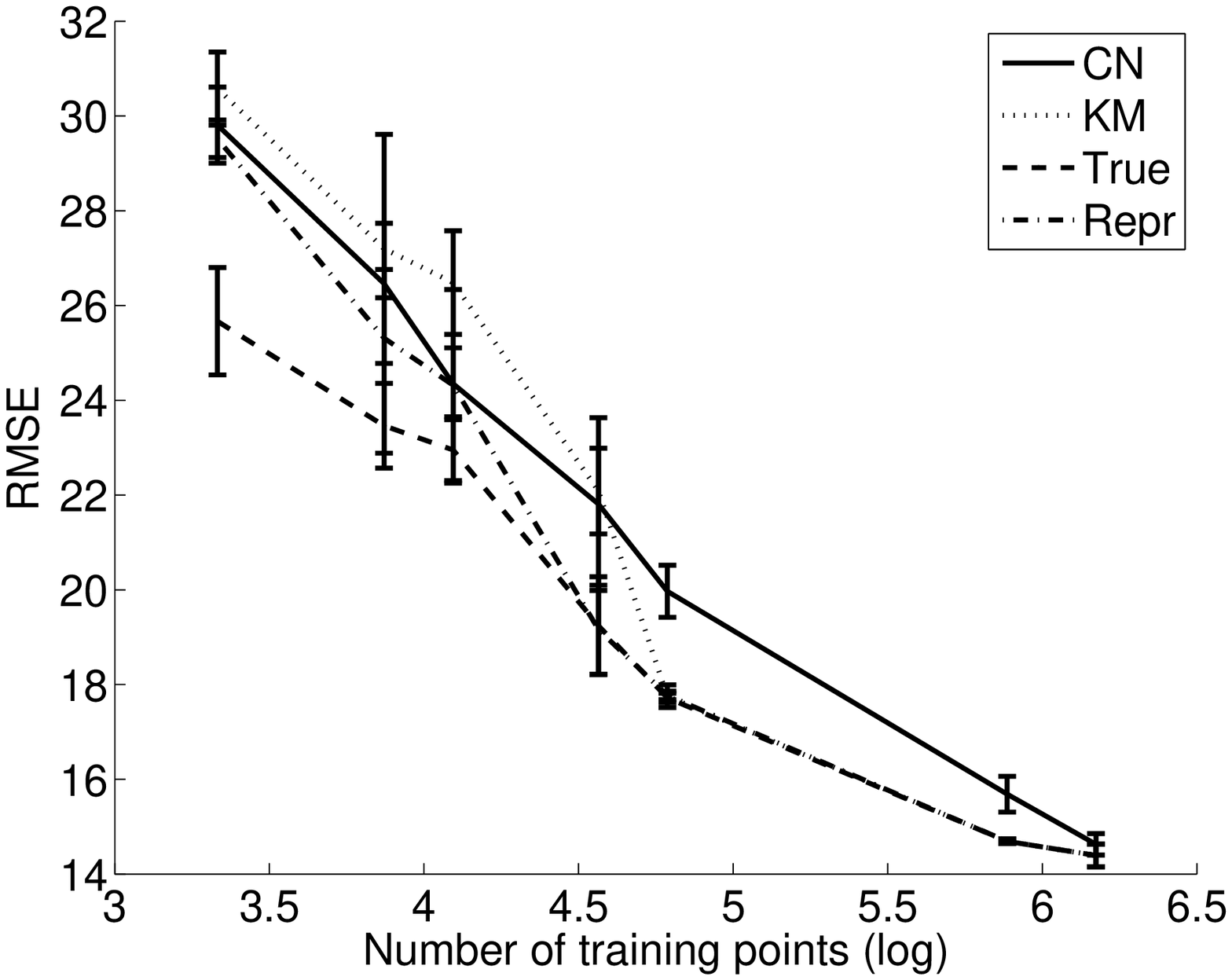}
\end{tabular}
\end{center}
\caption{RMSE versus number of training points for the tested methods.}
\label{fig:exp1}
\end{figure}

The results of the first experiment are shown on Figure~\ref{fig:exp1}
(left). As expected, both multi-task approaches perform better than
the approach that learns each task independently. CN penalization on
the other hand always gives better testing error than the trace norm
penalization, with a stronger advantage when very few training points
are available. When more training points become available, all the
methods give more and more similar performances. In particular, with
large samples, it is not useful anymore to use a multi-task approach.

\begin{figure}[h]
\begin{center}
\begin{tabular}{ccc}
  \includegraphics[width=.25\linewidth]{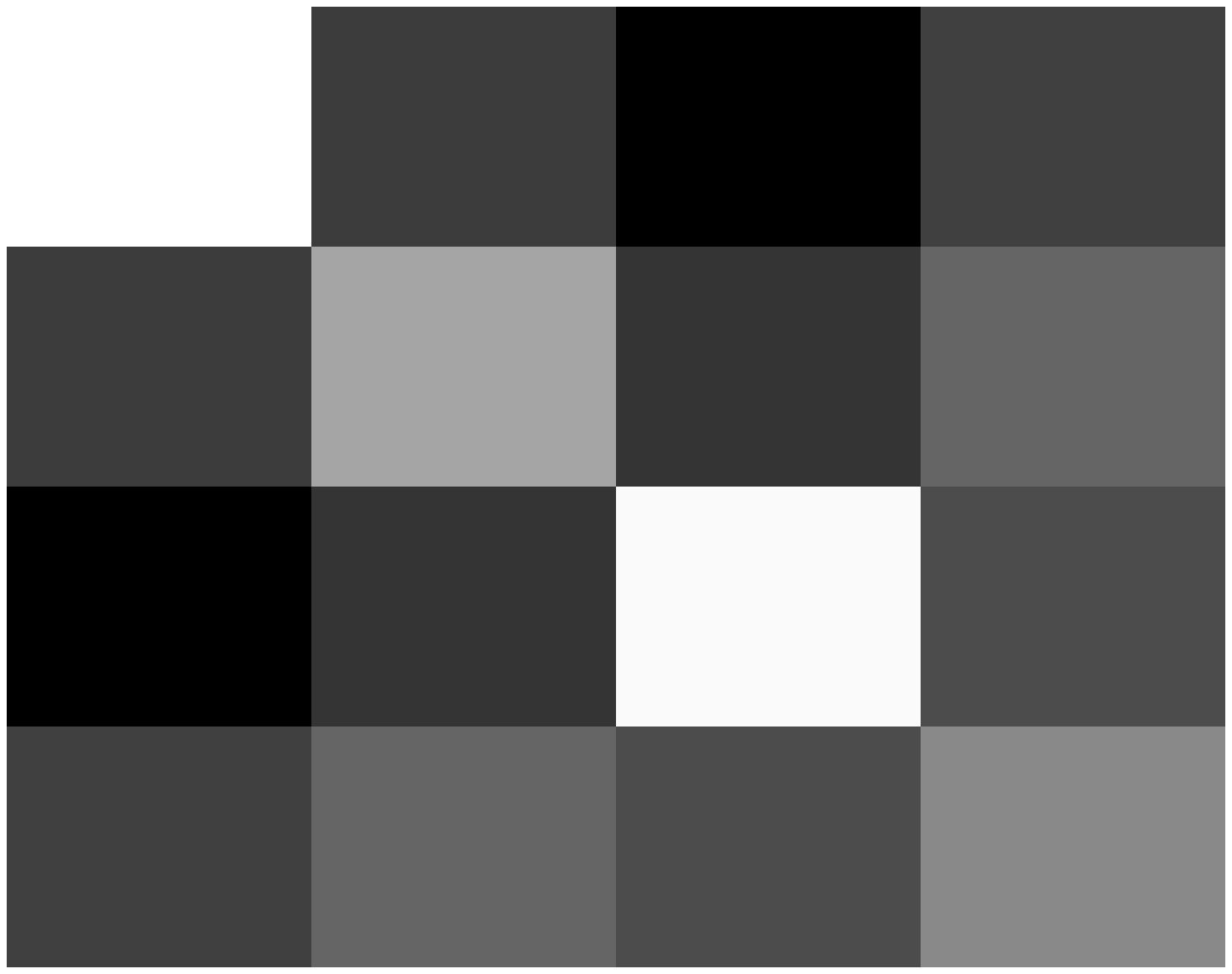} &
  \includegraphics[width=.25\linewidth]{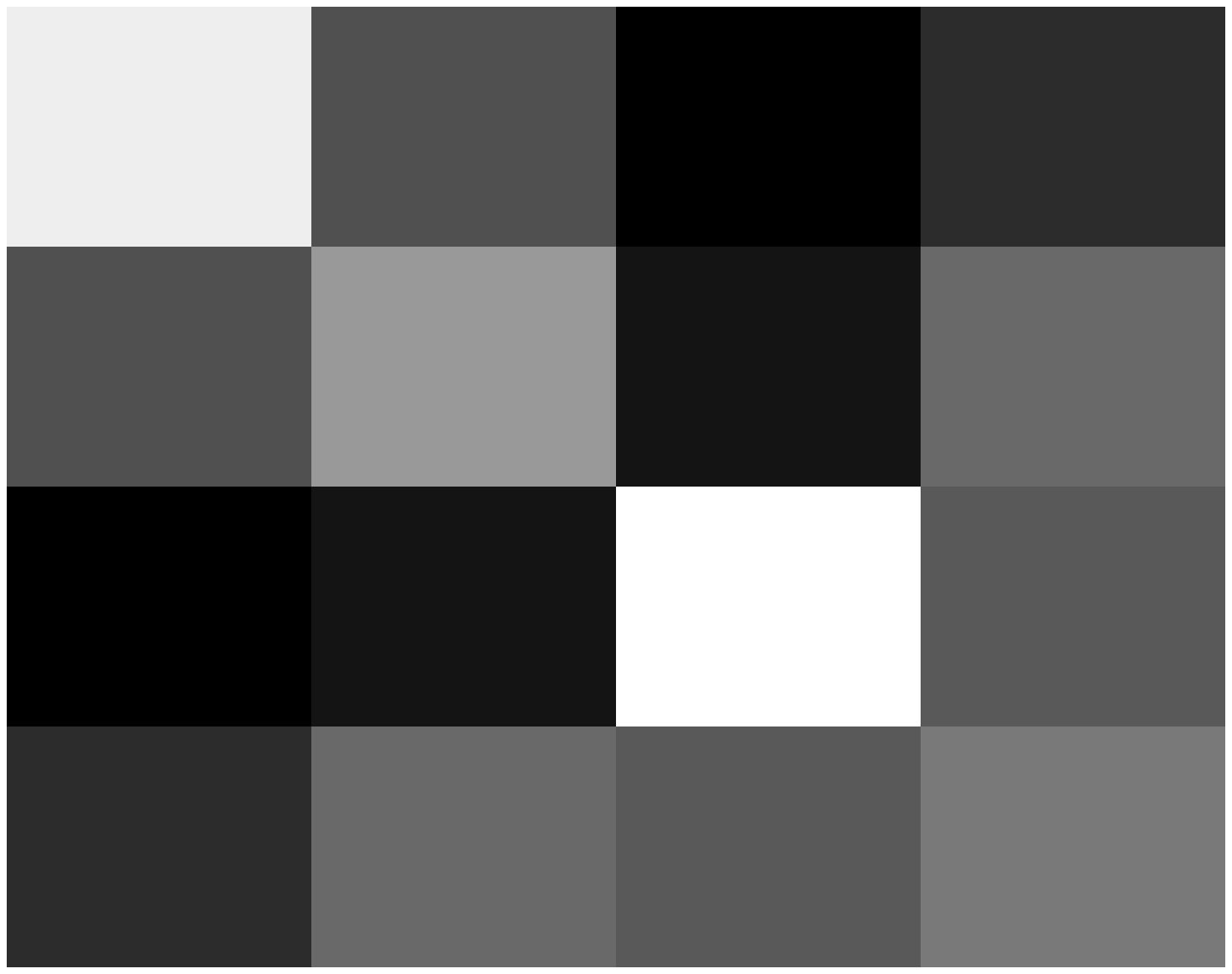} &
  \includegraphics[width=.25\linewidth]{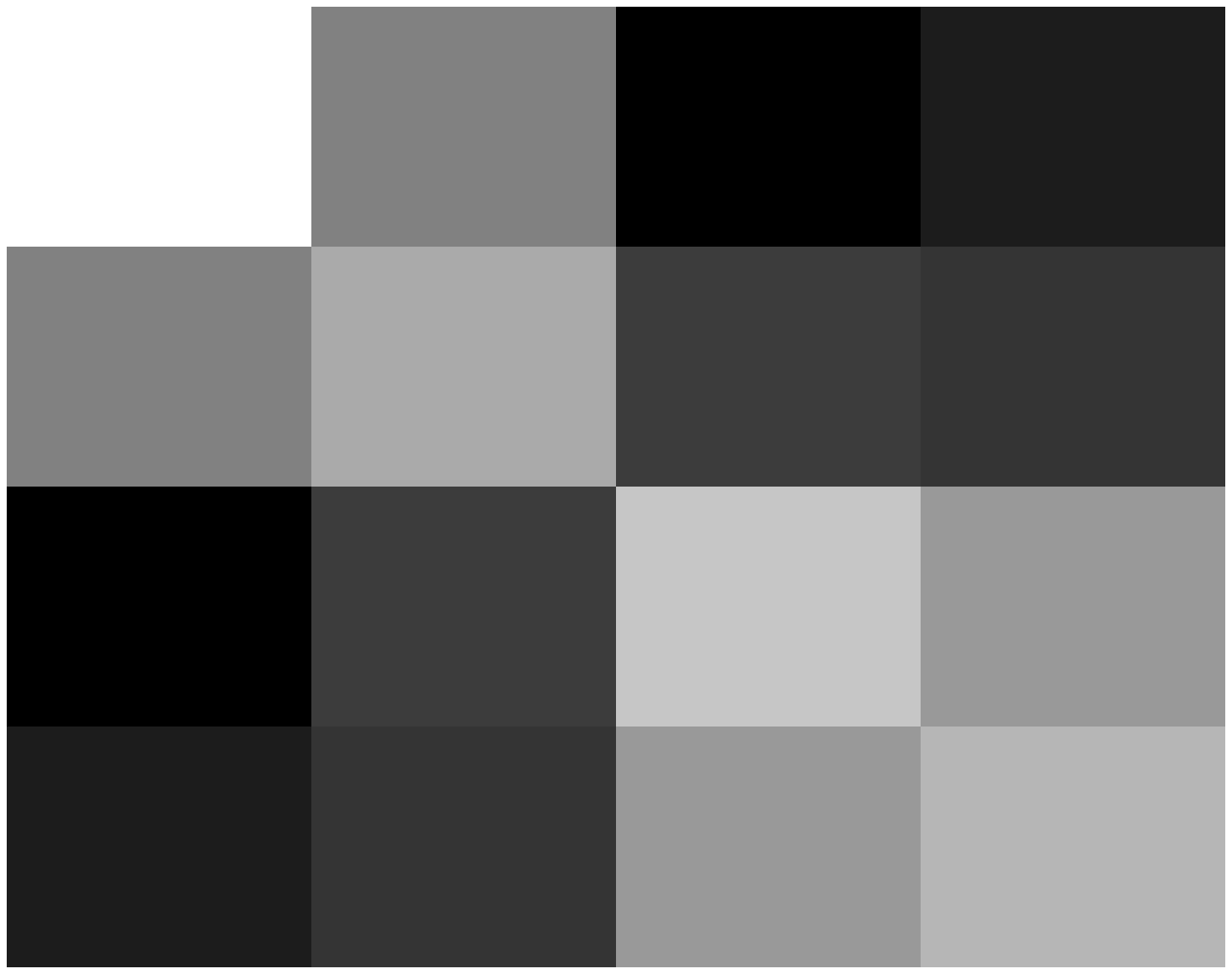} \\
  \includegraphics[width=.25\linewidth]{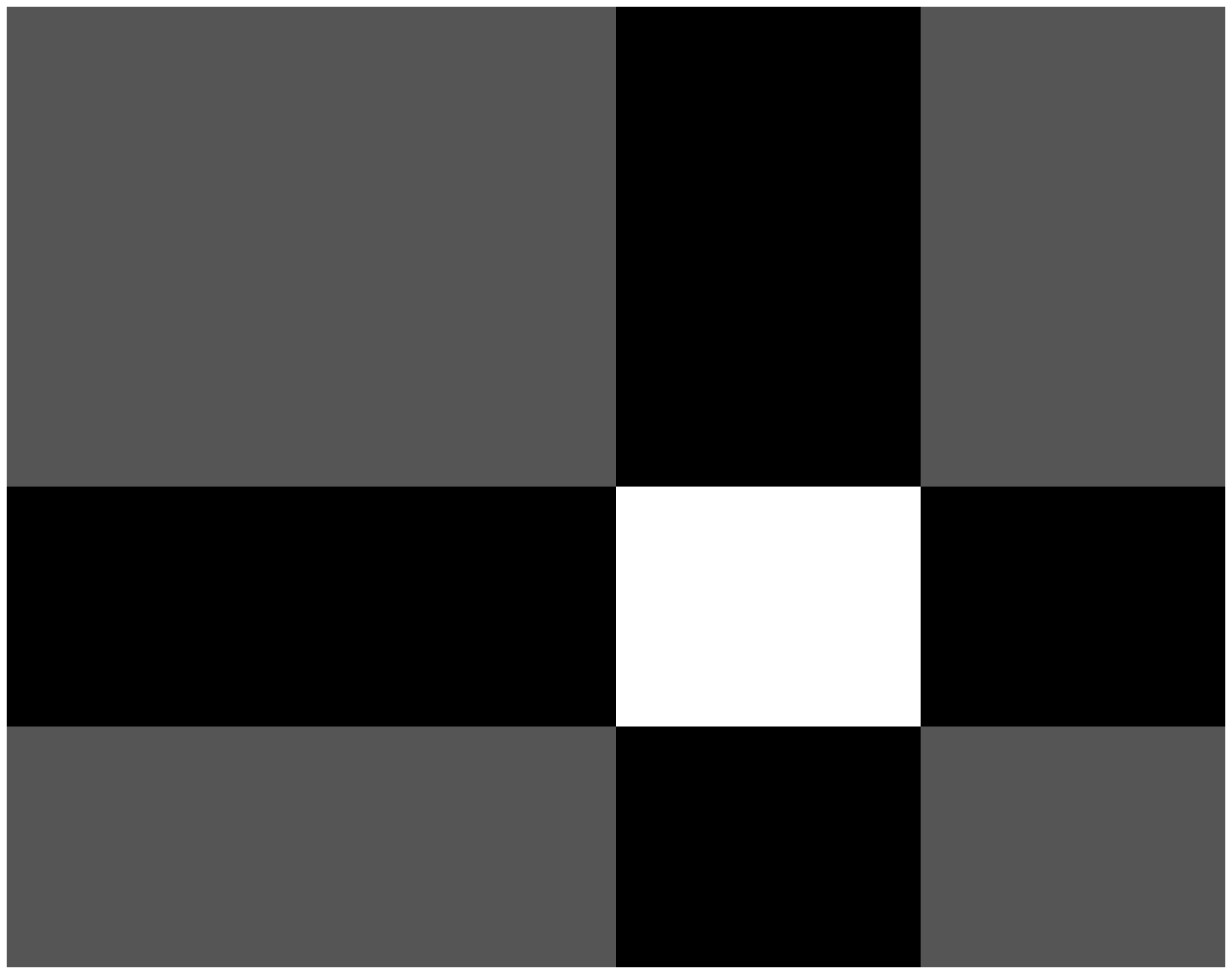} &
  \includegraphics[width=.25\linewidth]{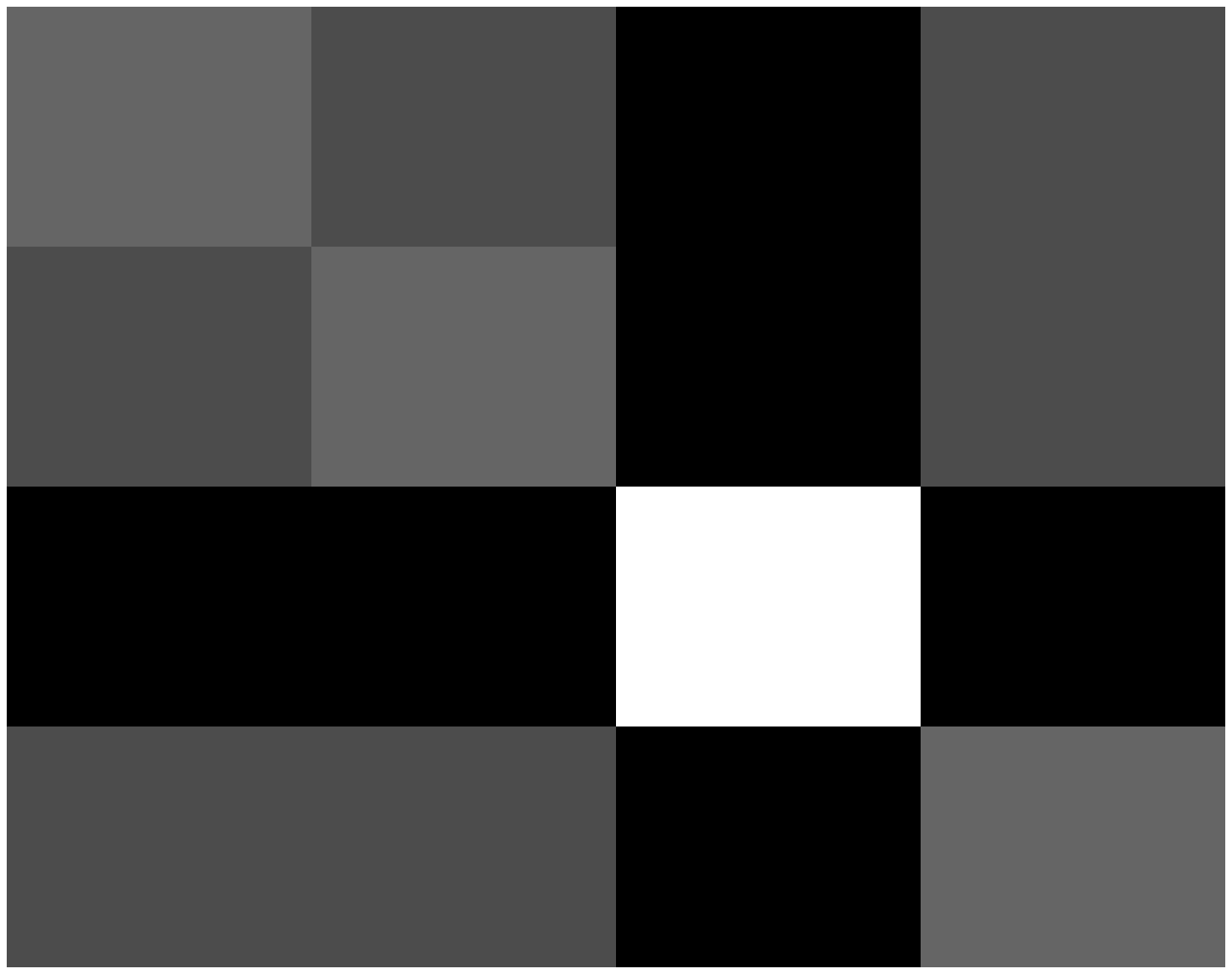} &
  \includegraphics[width=.25\linewidth]{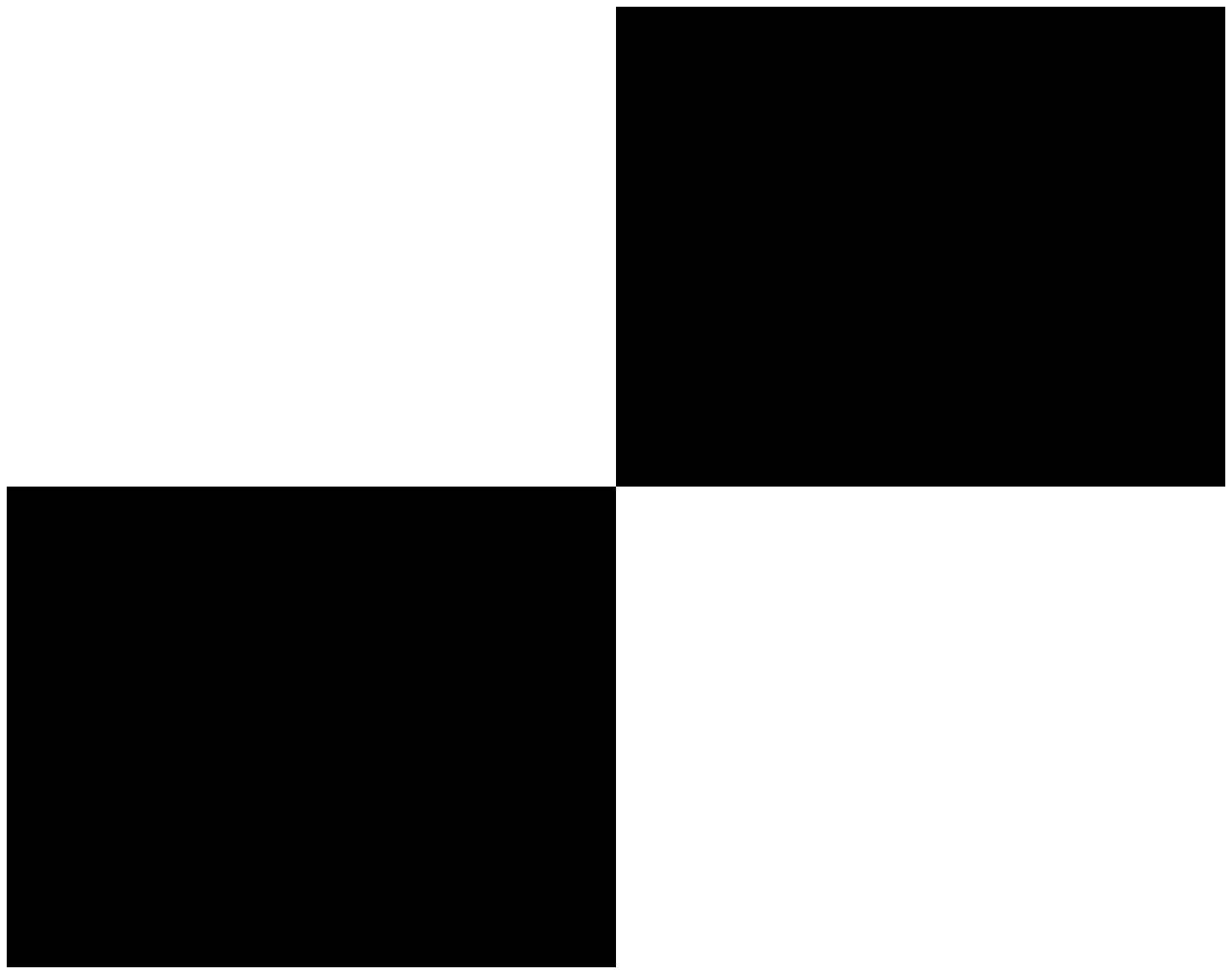}
\end{tabular}
\end{center}

\caption{Recovered $\Sigma$ with CN (upper line) and k-means (lower
  line) for $28$, $50$ and $100$ points.}
\label{fig:rsigmas}
\end{figure}

Figure~\ref{fig:exp1} (right) shows the results of the second
experiment.  Using the true metric always gives the best results. For
$28$ training points, no method recovers the correct clustering
structure, as displayed on Figure~\ref{fig:rsigmas}, although CN
performs slightly better than the k-means approach since the metric it
learns is more diffuse. For $50$ training points, CN performs much
better than the k-means approach, which completely fails to recover
the clustering structure as illustrated by the $\Sigma$ learned for
$28$ and $50$ training points on Figure~\ref{fig:rsigmas}.  In the
latter setting, CN partially recovers the clusters.  When more
training points become available, the k-means approach perfectly
recovers the clustering structure and outperforms the relaxed
approach. The reprojected approach, on the other hand, performs always
as well as the best of the two other methods. The CNinit approach
results are not displayed since the are the same as for the
reprojected method.

\subsection{MHC-I binding data}

We also applied our method to the \textsc{iedb} MHC-I peptide binding
benchmark proposed in~\cite{Peters2006community}. This database
contains binding affinities of various peptides, \emph{i.e.}, short
amino-acid sequences, with different MHC-I molecules. This binding
process is central in the immune system, and predicting it is crucial,
for example to design vaccines. The affinities are thresholded to give
a prediction problem. Each MHC-I molecule is considered as a task, and
the goal is to predict whether a peptide binds a molecule. We used an
orthogonal coding of the amino acids to represent the peptides and
balanced the data by keeping only one negative example for each
positive point, resulting in $15236$ points involving $35$ different
molecules. We chose a logistic loss for $\ell(W)$.

Multi-task learning approaches have already proved useful for this
problem, see for
example~\cite{Heckerman2006Leveraging,Jacob2008Efficient}. Besides, it
is well known in the vaccine design community that some molecules can
be grouped into empirically defined \emph{supertypes} known to have
similar binding behaviors.

\cite{Jacob2008Efficient} showed in particular that the multi-task
approaches were very useful for molecules with few known binders.
Following this observation, we consider the mean error on the $10$
molecules with less than $200$ known ligands, and report the results
in Table~\ref{tab:iedb}. We did not select the parameters by internal
cross validation, but chose them among a small set of values in order
to avoid overfitting. More accurate results could arise from such a
cross validation, in particular concerning the number of clusters
(here we limited the choice to $2$ or $10$ clusters).

\begin{table}[t]
  \caption{Prediction error for the $10$ molecules with less than $200$
    training peptides in \textsc{iedb}.}
\label{tab:iedb}
\begin{center}
\begin{tabular}{l c c c c c}
{\small\bf Method}  & {\small Pooling} & {\small Frobenius} &
{\small MT kernel} & {\small Trace norm} & {\small Cluster Norm}\\\hline
{\small\bf Test error} & {\small $26.53\% \pm 2.0$} & {\small $11.62\%
  \pm 1.4$} & {\small $10.10\% \pm 1.4$} & {\small $9.20\% \pm 1.3$} &
{\small $8.71\% \pm 1.5$}
\end{tabular}
\end{center}
\end{table}

The pooling approach simply considers one global prediction problem by
pooling together the data available for all molecules. The results
illustrate that it is better to consider individual models than one
unique pooled model, even when few data points are available. On the
other hand, all the multitask approaches improve the accuracy, the
cluster norm giving the best performance. The learned $\Sigma$,
however, did not recover the known supertypes, although it may contain
some relevant information on the binding behavior of the molecules.
Finally, the reprojection methods (\emph{reprojected} and
\emph{CNinit}) did not improve the performance,
potentially because the learned structure was not strong enough.

\section{Conclusion}

We have presented a convex approach to clustered multi-task learning,
based on the design of a dedicated norm. Promising results were
presented on synthetic examples and on the \textsc{iedb} dataset. We
are currently investigating more refined convex relaxations and the
natural extension to non-linear multi-task learning as well as the
inclusion of specific features on the tasks, which has shown to
improve performance in other settings~\cite{Abernethy2006Low-rank}.

\bibliographystyle{unsrt} 
\bibliography{../../../bibli/bibli}

\end{document}